# Can AI Make Energy Retrofit Decisions? An Evaluation of Large Language Models

Lei Shu [a,b*] and Dong Zhao [a,b,c]

a. School of Planning, Design, and Construction, Michigan State University, East Lansing, Michigan, USA.
b. Human-Building Systems Lab, Michigan State University, East Lansing, Michigan, USA
c. Department of Civil and Environmental Engineering, Michigan State University, East Lansing, Michigan, USA.

* Corresponding author: Lei Shu, School of Planning, Design & Construction, Michigan State University, 552 W Circle Dr, East Lansing, Michigan, United States. Email: shulei1@msu.edu

## Abstract

Conventional approaches to building energy retrofit decision-making face significant challenges, including poor generalizability and low interpretability. These issues hinder their practical adoption, particularly in diverse residential contexts. Recent advancements in Smart and Connected Communities suggest that generative AI, specifically large language models (LLMs), may help overcome these limitations by processing complex contextual information and generating interpretable, human-readable retrofit recommendations. This study evaluates the capabilities of seven leading LLMs—ChatGPT, DeepSeek, Gemini, Grok, Llama, and Claude—in generating retrofit decisions under two contexts: maximizing $CO_2$ reduction (technical) and minimizing payback period (sociotechnical). Model performance is assessed along four dimensions: accuracy, consistency, sensitivity, and reasoning, using a dataset of 400 diverse homes across 49 U.S. states. Results show that LLMs can generate effective retrofit recommendations, with accuracy reaching up to 54.5% for top-1 matches and 92.8% within the top-5, despite no model fine-tuning. The performance is notably stronger in the technical context, reflecting the models' ability to optimize clear engineering objectives, while performance in the sociotechnical context is limited by trade-offs in economic and contextual considerations. Consistency among LLMs is low, with higher-performing models trending to diverge from others. LLMs are highly sensitive to the geographical location and space geometry of buildings while less sensitive to their technologies and occupant behavior. Most LLMs use structured, step-by-step reasoning that aligns with engineering-style logic; however, this approach is often simplified and lacks deeper contextual awareness or nuanced understanding. Overall, LLMs demonstrate promising capabilities in making energy retrofit decisions, although improvements in accuracy, consistency, and contextual understanding are necessary for reliable application in practice.

# 1. Introduction

Building energy consumption is governed by physical building characteristics, climate conditions, and occupant behaviors [1]. Building structural and thermal properties set a baseline for energy use, climate conditions drive heating and cooling loads, and occupant behaviors can dramatically sway actual consumption. These factors not only determine energy consumption but also critically influence energy retrofit performance. For instance, adding insulation to exterior walls reduces heat loss in buildings with poor insulation but provides only marginal benefits in buildings that already have moderate insulation. A cool roof retrofit helps lower indoor temperatures and reduce cooling demand in hot climates but has little effect in colder regions where heating demands dominate over cooling. Meanwhile, occupant behaviors like adjusting thermostat settings can counteract the expected energy savings of a heating, ventilation, and air conditioning (HVAC) upgrade—for example, setting the heating temperature higher in winter or the cooling temperature lower in summer than originally assumed.

Existing decision-making methods for building retrofits rely on either physics-based or data-driven methods [2, 3], however, they have inherent limitations in handling the aforementioned three factors. Physics-based methods rely on heat and mass transfer principles as well as energy balance to simulate building energy performance and retrofit outcomes. They are physically grounded and known for their high modeling accuracy, making them a widely accepted reference in performance evaluation studies and engineering practice [4, 5]. However, they face several challenges, including heavy building input, limited scalability, and static behavior assumptions. First, physics-based simulation require technically detailed and complex input data [5]. Building characteristics, HVAC system efficiency, and domestic hot water system efficiency can typically be obtained from building audits, blueprints, or energy performance certificates [6, 7]. However, reliably collecting these parameters and accurately inputting them into building energy simulations become increasingly complex and often impractical as the number of buildings grows, posing a significant barrier to large-scale applications [3]. Second, these methods exhibit limited scalability not only due to their high data requirements, but also because they often overlook between-building interactions, such as shared energy infrastructure, district systems, and multi-stakeholder coordination [8]. Even when effective at the individual building level, extending them to the community or city scale remains challenging without system-level integration. Third, climate conditions, such as outdoor temperature, solar radiation, and humidity levels, are commonly obtained from empirical meteorological data of reference cities within the same climate zone, for instance, those provided by the EnergyPlus Weather Data Sources [9]. However, low climate sensitivity to local microclimatic variations—such as urban heat island effects—can lead to discrepancies between simulated and actual building performance [10]. Finally, occupant behavior, such as thermostat settings, appliance usage patterns, and lighting setups, are often represented using static schedules and rules, overlooking dynamic interactions and behavioral uncertainty. This static behavior assumption can introduce significant discrepancies in energy predictions, particularly for small-scale buildings, where a smaller sample size magnifies the relative error between simulated and actual energy use [11].

Data-driven methods rely on historical energy data, statistical models, and machine learning algorithms to assess and predict the performance of retrofit measures [3]. These methods offer empirical learning and flexible modeling, making them well-suited for diverse datasets. However, they also face distinct challenges, including the lack of baseline data, limited generalizability, and the limited interpretability. First, the lack of baseline data is particularly evident in older renovated buildings, which often do not have comprehensive records of pre-retrofit and post-retrofit energy performance. Even if upgrades such as improved insulation meet current building codes, the absence of prior data makes it difficult to accurately assess actual energy savings [12]. Second, ensuring generalizability is challenging [13]. For example, a machine learning model trained in cold climates may not work well in a hot climate, where buildings have different energy use patterns and may respond differently to the same retrofit measures. Finally, many data-driven models suffer from limited interpretability. Their underlying mechanisms are often opaque, making it difficult for stakeholders—especially non-experts—to understand how predictions are generated or to trust the resulting recommendations [14, 15].

Smart and Connected Communities (S&CCs) offer a promising environment for addressing key challenges in building energy retrofits by integrating a suite of emerging technologies such as Digital Twins, Multi-Agent Systems (MAS), and Artificial Intelligence (AI), as visually summarized in Figure 1. Digital Twins, enabled by a combination of technologies—including Building Information Modeling (BIM) and the Internet of Things (IoT)—provide real-time, high-resolution data on building operations, local climate, and occupant behavior. This reduces reliance on heavy building input, improves responsiveness to microclimatic variations, and overcomes static behavior assumptions [16, 17]. It also helps mitigate missing baseline data by generating continuous records of building performance [18]. MAS, encompassing Energy System Modeling and Social Network Modeling, enables simulation of interactions across buildings and infrastructures. These models address limited scalability in physics-based approaches by capturing between-building dynamics and supporting coordinated decision-making at the large scale [8]. Recent generative AI techniques, like large language models (LLMs), improve generalizability by learning contextual patterns from vast datasets [19], enabling performance predictions across varied building types and climates. Additionally, their natural language capabilities offer new pathways for improving interpretability, by translating complex model outputs into human-readable explanations [20].

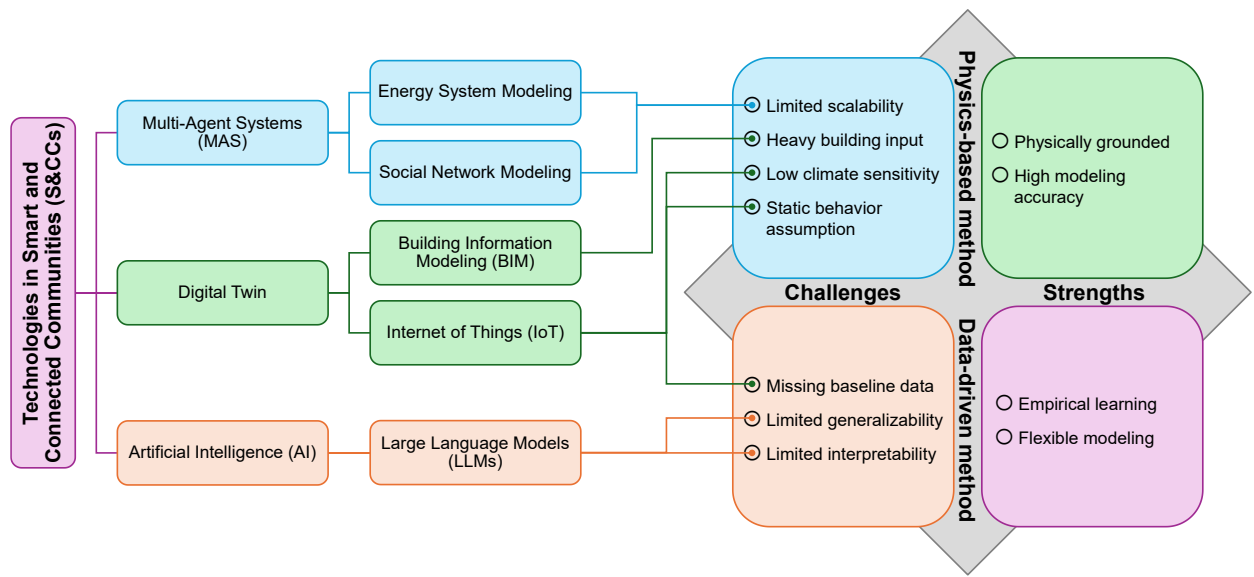

Figure 1. Smart and Connected Communities (S&CCs) technologies for overcoming challenges in decision-making methods for building retrofits.

The recent advances in LLM-based AI can overcome the "last-mile" challenge by translating and synthesizing complex technical data into clear, actionable, and personalized recommendations [21, 22]. In the context of an S&CC ecosystem, LLMs act as conversational interfaces that connect expert knowledge with everyday decision-making, fostering greater engagement among homeowners, building managers, and community members [23, 24]. For instance, they can offer tailored retrofit advice—such as insulation upgrades, HVAC improvements, or solar system integration—while simultaneously factoring in occupant comfort, budget constraints, and climate conditions [25]. Homeowners can interactively explore various retrofit options using natural language, receiving explanations about potential energy savings, payback periods, and impacts on indoor comfort [26]. While the emerging applications demonstrate the promise of LLM-based AI in supporting retrofit decisions, most models are trained on broad, general-purpose datasets rather than domain-specific knowledge. As a result, their capability to generate reliable energy retrofit recommendations that reflect diverse building characteristics, climate conditions, and occupant behaviors remains unclear. The lack of systematic evaluation further raises concerns about the practical reliability of their outputs in practical retrofit applications [22, 27].

To address the lack of domain-specific evaluation, the objective of this study is to evaluate the performance of LLMs in residential building energy retrofit decision-making. Specifically, we examine their ability to identify optimal retrofit options based on environmental benefit and economic feasibility. We selected seven widely discussed LLMs and LLM-powered applications—ChatGPT o1, ChatGPT o3, DeepSeek R1, Grok 3, Gemini 2.0, Llama 3.2, and Claude 3.7—released between September 2024 and February 2025. OpenAI's ChatGPT o1 [28] and o3 [29] are recognized for their balanced reasoning and coherent multi-step responses. DeepSeek R1 from DeepSeek delivers top-tier performance in coding and math while remaining highly affordable for real-world deployment [30]. Grok 3 from xAI is uniquely integrated with real-time data from X

(formerly Twitter), enabling up-to-date responses [31]. Gemini 2.0 from Google DeepMind is notable for its strong multimodal capabilities, particularly in handling images and long documents [32]. Llama 3.2 from Meta is distinguished by its lightweight, open-access design that supports flexible fine-tuning [33]. Claude 3.7 from Anthropic is a hybrid reasoning model that combines fast and deep thinking with top-tier coding, reliable instruction-following, and advanced data understanding [34]. For simplicity, these LLMs and LLM-powered applications are hereafter collectively referred to as LLMs.

The performance of these LLMs was evaluated across four key dimensions: accuracy in matching established baseline, consistency across models, sensitivity to contextual features (including building characteristics, climate conditions, and occupant behaviors), and the quality of their reasoning. This multi-faceted evaluation provides a foundation for understanding the strengths and limitations of the current LLM-based AI in making retrofit decisions throughout diverse residential contexts.

## 2. Methodology

2.1. Data

This study used a selected subset of the ResStock 2024.2 dataset from the National Renewable Energy Laboratory (NREL), containing 550,000 representative residential building samples [35]. For our analysis, we selected 400 homes from 49 U.S. states (excluding Hawaii). These homes represent a variety of building types, including single-family homes, multi-family residences, and mobile homes. They were constructed over a wide time span, ranging from buildings built before 1940 to those built around 2010. The floor area of these buildings varies from 25 to 519 $m^2$.

Each sample is defined by 389 parameters, covering building properties (e.g., construction vintage, floor area, insulation levels), equipment characteristics (e.g., HVAC system type and efficiency, appliance ownership), occupant attributes (e.g., household size, usage patterns, income), location details (e.g., postal code, census region, state), and energy usage (e.g., utility bill rate, energy consumption, energy cost). These data primarily come from the U.S. Energy Information Administration's (EIA) Residential Energy Consumption Survey (RECS), the U.S. Census Bureau's American Housing Survey (AHS), and American Community Survey (ACS) [35].

The selected dataset includes parameters for retrofit measures (e.g., infiltration and insulation upgrades, heat pump upgrades, appliance electrification) and energy-related outputs (e.g., energy consumption, $CO_2$ emissions, energy costs). Specifically, the dataset provides 16 retrofit packages. The first 15 vary along three main axes: the type and efficiency of the heat pump, whether infiltration and attic-insulation upgrades are applied, and whether major appliances are electrified. The 16th package solely applies infiltration and insulation upgrades. Thus, each building sample features 17 sets of energy-related outputs: one baseline scenario with no retrofits and 16 retrofit scenarios (one per package). All energy-related outputs are computed by EnergyPlus simulations [35].

The ResStock 2024.2 dataset does not include cost data for the retrofit measures, we supplemented our dataset with unit price estimates from the National Residential Efficiency Measures (NREM) Database [36]. These unit prices were then combined with building-specific parameters—such as floor area and equipment capacity—to calculate the total cost of each retrofit. Table 1 provides the technical details and estimated costs for the 16 retrofit packages, converted to International System (SI) units where appropriate.

Table 1. Technical specifications and cost estimates for 16 retrofit packages.

| No. | Heat Pump (Cost) | Envelope (Cost) | Appliances (Cost) |
|---|---|---|---|
| 1 | ENERGY STAR ASHP (SEER 16, HSPF 9.2) with electric backup ($3,700 + $143 × capacity [kW]) | None | None |
| 2 | High-efficiency cold-climate ASHP (SEER 20, HSPF 11) with electric backup ($4,000 + $143 × capacity [kW]) | None | None |
| 3 | Ultra high-efficiency ASHP (SEER 24, HSPF 13) with electric backup ($4,600 + $143 × capacity [kW]) | None | None |
| 4 | ENERGY STAR dual-fuel ASHP (SEER 16, HSPF 9.2) with existing system as backup ($3,700 + $143 × capacity [kW]) | None | None |
| 5 | ENERGY STAR geothermal HP (EER 20.5, COP 4.0) ($870 + $887 × capacity [kW]) | None | None |
| 6 | ENERGY STAR ASHP (SEER 16, HSPF 9.2) with electric backup ($3,700 + $143 × capacity [kW]) | Infiltration: 30% reduction ($3.34–$12.92 / m²) Ceiling insulation: RSI 8.6–10.6 ($5.06–$29.06 / m²) | None |
| 7 | High-efficiency cold-climate ASHP (SEER 20, HSPF 11) with electric backup ($4,000 + $143 × capacity [kW]) | Infiltration: 30% reduction ($3.34–$12.92 / m²) Ceiling insulation: RSI 8.6–10.6 ($5.06–$29.06 / m²) | None |
| 8 | Ultra high-efficiency ASHP (SEER 24, HSPF 13) with electric backup ($4,600 + $143 × capacity [kW]) | Infiltration: 30% reduction ($3.34–$12.92 / m²) Ceiling insulation: RSI 8.6–10.6 ($5.06–$29.06 / m²) | None |
| 9 | Dual-fuel ASHP (SEER 16, HSPF 9.2) with existing system as backup ($3,700 + $143 × capacity [kW]) | Infiltration: 30% reduction ($3.34–$12.92 / m²) Ceiling insulation: RSI 8.6–10.6 ($5.06–$29.06 / m²) | None |
| 10 | ENERGY STAR geothermal HP (EER 20.5, COP 4.0) ($870 + $887 × capacity [kW]) | Infiltration: 30% reduction ($3.34–$12.92 / m²) Ceiling insulation: RSI 8.6–10.6 ($5.06–$29.06 / m²) | None |
| 11 | ENERGY STAR ASHP (SEER 16, HSPF 9.2) with electric backup ($3,700 + $143 × capacity [kW]) | Infiltration: 30% reduction ($3.34–$12.92 / m²) Ceiling insulation: RSI 8.6–10.6 ($5.06–$29.06 / m²) | Dryer ($760), Induction Range ($1,900), HPWH ($2,000–$2,300) |
| 12 | High-efficiency cold-climate ASHP (SEER 20, HSPF 11) with electric backup ($4,000 + $143 × capacity [kW]) | Infiltration: 30% reduction ($3.34–$12.92 / m²) Ceiling insulation: RSI 8.6–10.6 ($5.06–$29.06 / m²) | Dryer ($760), Induction Range ($1,900), HPWH ($2,000–$2,300) |
| 13 | Ultra high-efficiency ASHP (SEER 24, HSPF 13) with electric backup ($4,600 + $143 × capacity [kW]) | Infiltration: 30% reduction ($3.34–$12.92 / m²) Ceiling insulation: RSI 8.6–10.6 ($5.06–$29.06 / m²) | Dryer ($760), Induction Range ($1,900), HPWH ($2,000–$2,300) |
| 14 | Dual-fuel ASHP (SEER 16, HSPF 9.2) with existing system as backup ($3,700 + $143 × capacity [kW]) | Infiltration: 30% reduction ($3.34–$12.92 / m²) Ceiling insulation: RSI 8.6–10.6 ($5.06–$29.06 / m²) | Dryer ($760), Induction Range ($1,900), HPWH ($2,000–$2,300) |

| 15 | ENERGY STAR geothermal HP (EER 20.5, COP 4.0) ($870 + $887 × capacity [kW]) | Infiltration: 30% reduction ($3.34–$12.92 / m²) Ceiling insulation: RSI 8.6–10.6 ($5.06–$29.06 / m²) | Dryer ($760), Induction Range ($1,900), HPWH ($2,000–$2,300) |
|---|---|---|---|
| 16 | No heat pump change | Infiltration: 30% reduction ($3.34–$12.92 / m²) Ceiling insulation: RSI 8.6–10.6 ($5.06–$29.06 / m²) | None |

**Abbreviations:** ASHP: Air-Source Heat Pump; SEER: Seasonal Energy Efficiency Ratio; HSPF: Heating Seasonal Performance Factor; EER: Energy Efficiency Ratio; COP: Coefficient of Performance; RSI: Thermal Resistance in SI units; HPWH: Heat Pump Water Heater.

2.2. Prompt

As shown in Figure 2, the prompt comprises three parts: an overview of the 16 retrofit packages, an assigned role and question, and house-specific information. The overview describes each measure's features (e.g., heat pump efficiency, whether infiltration is upgraded, whether major appliances are electrified) along with its associated costs. The assigned role places the LLM in the position of a "house retrofit specialist," tasked with evaluating multiple homes, comparing costs and efficiency across all packages, and identifying which retrofit delivers the greatest $CO_2$ reduction and which provides the lowest payback period for each home.

The house-specific information is drawn from 28 key parameters, screened from the original 389 in ResStock 2024.2, focusing on factors essential to energy consumption and readily known by most households. These parameters encompass location information (e.g., county name and state name), architectural features (e.g., floor area and orientation), building envelope details (e.g., insulation levels and window type), appliance types (e.g., clothes dryer type and lighting type), water heating systems (e.g., water heater efficiency and type), HVAC systems (e.g., heating fuel and space cooling type), and occupant-related usage patterns (e.g., indoor temperature setpoint and occupant number). Rather than relying on more technical metrics—such as the climate zone, which are generally unfamiliar to most households—this selection prioritizes parameters that are commonly understood. The included location information, however, still allows the LLM to infer local climate characteristics. Since the dataset was originally structured in a technical, tabular format, we used GPT-4o to reframe each entry from households' perspective, aligning the descriptions with everyday knowledge while retaining crucial details for evaluating retrofit measures.

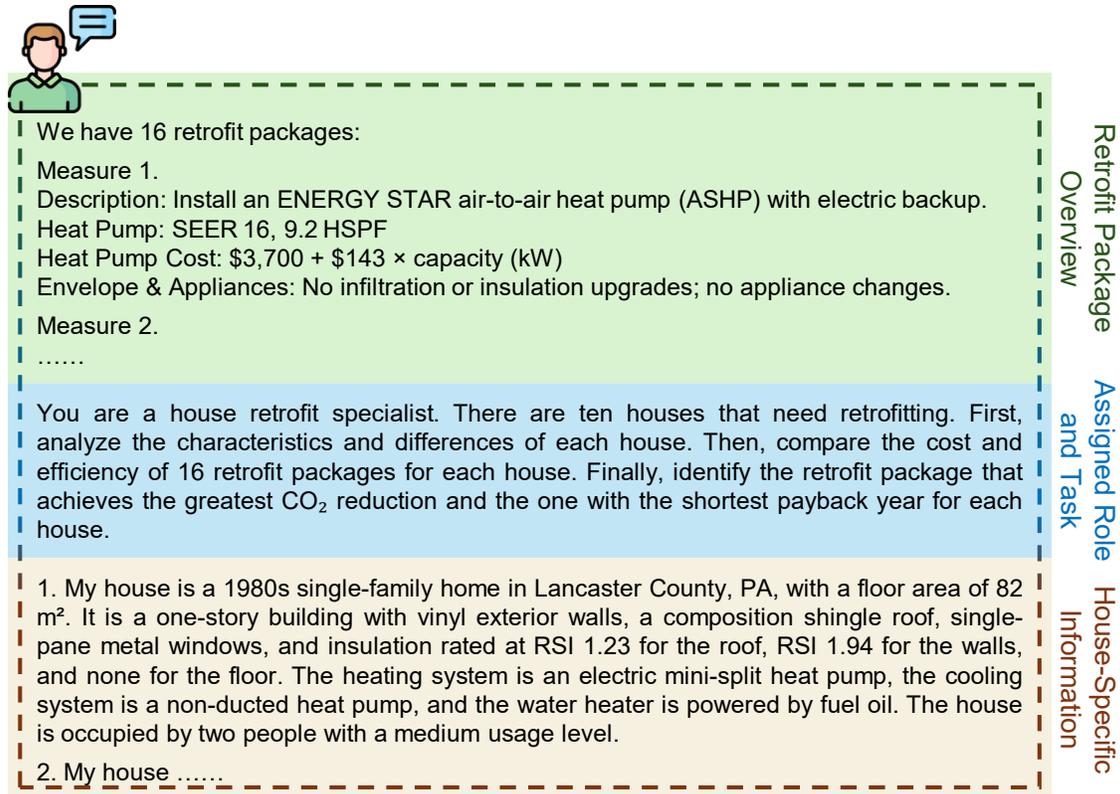

Figure 2. Prompt structure for LLMs retrofit decision-making.

2.3. Evaluation

We evaluated seven LLMs—ChatGPT o1, ChatGPT o3, DeepSeek R1, Grok 3, Gemini 2.0, Llama 3.2, and Claude 3.7. For each sample house, we identified two baseline retrofit measures: one maximizing annual $CO_2$ reduction, and one minimizing payback period. The $CO_2$ reduction-driven measure was identified directly from the EnergyPlus simulation outputs provided in the dataset. The payback period-driven measure was determined by dividing the total retrofit cost by the annual energy cost savings. The energy savings were obtained from the dataset, and the total retrofit costs were calculated as described previously using unit prices from the NREM Database and building-specific parameters.

Although the dataset initially featured 16 retrofit packages, three of the five heat pumps differed only in nominal efficiency while retaining the same air-to-air design and backup system. To streamline evaluation without sacrificing interpretability, we grouped similar packages into broader categories, reducing the set from 16 to 10 consolidated options. This consolidation allowed for a more concise analysis without meaningfully altering the breadth of retrofit options.

Each LLM was instructed to recommend a retrofit package that either maximizes $CO_2$ reduction or minimizes payback period. We then mapped each selection to its consolidated category and evaluated how closely it aligned with the baseline. Because retrofit decisions often involve multiple near-optimal options, we introduced a tiered evaluation to reflect different levels of alignment. A selection was classified as Top-1 if it exactly matched the top-ranked measure, Top-3 if it fell within the top three, and Top-5 if it appeared among the top five.

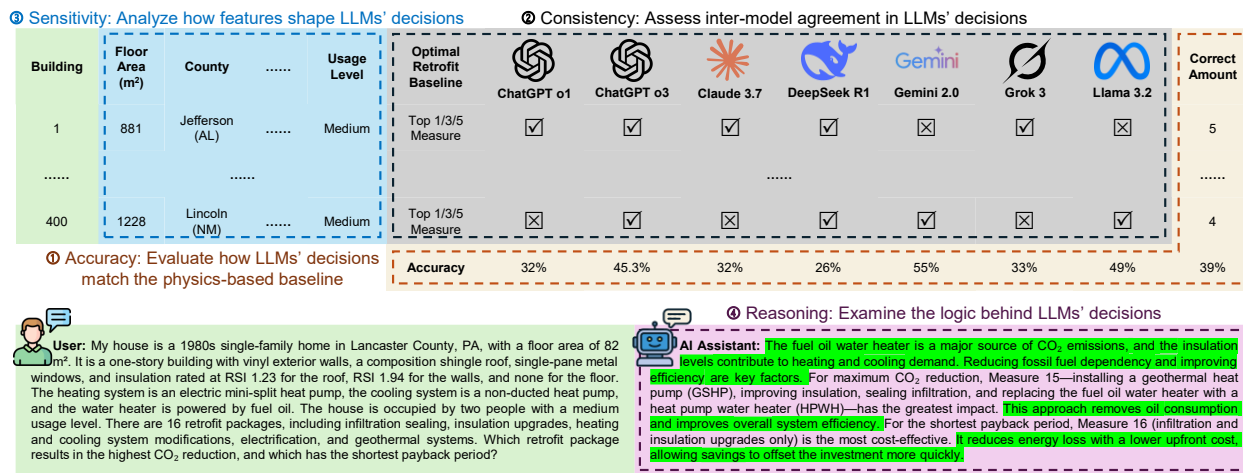

Figure 3. Four-dimensional evaluation of LLM performance in retrofit decision-making.

Beginning with the four-dimension evaluation (Figure 3), we evaluated accuracy to measure how closely each LLM's selection aligned with the baseline. Accuracy was assessed at three tiers of decision criteria: Top-1, Top-3, and Top-5. An LLM's selection was considered accurate if the recommended retrofit package was among the top-$k$ options identified by the physics-based model, where $k$ = 1, 3, or 5. For example, if the baseline model ranked options A, B, and C as the top 3, and the LLM recommended option B, it would not be considered accurate for Top-1, but would be considered accurate for Top-3 and Top-5. Accuracy was calculated using the following formula:

$$accuracy = \frac{c}{n} \quad (1)$$

where, $c$ is the number of correct selection and $n$ is the total number of evaluated cases. This analysis was conducted separately for two objectives: maximizing $CO_2$ reduction and minimizing payback year.

Second, we evaluated consistency, which captures how reliably the LLMs agree with one another, revealing whether they tend to converge on similar choices or follow divergent heuristics. Three inter-rater reliability metrics were used: Fleiss' Kappa, leave-one-out Fleiss' Kappa, and pairwise Cohen's Kappa. All three are derived from the following general formulation for computing the Kappa value:

$$kappa = \frac{p_o - p_e}{1 - p_e} \quad (2)$$

where $p_o$ represents the observed agreement, and $p_e$ is the agreement expected by chance. Fleiss' Kappa measures overall agreement across all models. Leave-one-out Fleiss' Kappa measures agreement after excluding one model at a time to assess its contribution to overall agreement. Cohen's Kappa measures pairwise agreement between every model pair. These metrics were calculated under both $CO_2$ reduction and payback year objectives, and for Top-1, Top-3, and Top-5 levels. In each case, we examined both selection agreement (whether models selected the same retrofit) and correctness agreement (whether they made the same correct or incorrect judgment relative to the baseline).

Third, we analyzed model sensitivity to input features to assess whether LLMs prioritize features similarly to physics-based baselines. Using a random forest classifier, we treated the selected retrofit option from each LLM and the baseline as the target variable, and the 28 parameters as predictors. The resulting percentage feature importance indicated the relative influence of each parameter on the model's decisions. This analysis was performed separately for the $CO_2$ reduction and payback year objectives to compare feature reliance across models. Alignment with the physics-based baseline suggests that LLMs may be weighing inputs in a domain-consistent manner.

Fourth, we examined the reasoning processes underlying LLMs' retrofit decisions to evaluate whether models can generate outputs through physically grounded logic. Among the seven models, only ChatGPT o3 and DeepSeek R1 provided detailed step-by-step reasoning. Focusing on these two, we conducted a qualitative analysis, offering insight into the internal logic guiding their decisions.

3. Results

3.1. Accuracy

Figure 4 shows the Top-1 (blue bar), Top-3 (orange bar), and Top-5 (green bar) accuracy with 95% confidence intervals for seven LLMs and their overall performance in selecting optimal retrofit measures. The top chart shows accuracy for selecting measures that maximize $CO_2$ reduction, while the bottom chart shows accuracy for those that minimize payback year. The "Overall" bars, shown alongside individual models, represent the average accuracy across all LLMs for each objective.

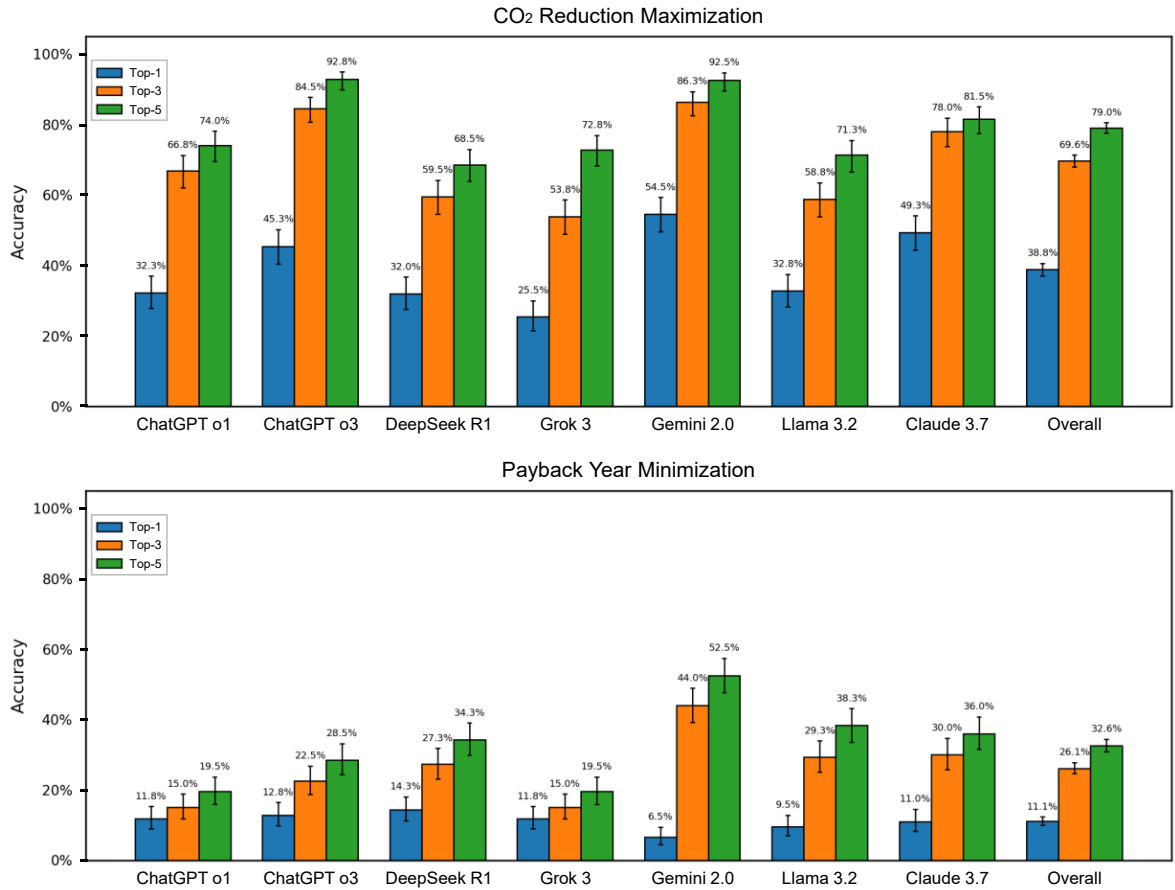

Figure 4. Top-$k$ accuracy (with 95% CI) of LLMs in selecting optimal retrofit measures.

For performance across objectives, LLMs achieve higher accuracy for $CO_2$ reduction than for payback year, and accuracy improves from Top-1 to Top-5. In the $CO_2$ task, overall performance reaches fair accuracy at Top-3 and good accuracy at Top-5, while payback year accuracy remains poor throughout. Specifically, the average Top-1 accuracy is 38.8% for $CO_2$ reduction compared to 11.1% for payback year; at Top-3, 69.6% versus 26.1%; and at Top-5, 79.0% versus 32.6%.

For $CO_2$ reduction maximization, the stronger-performing LLMs achieve marginal Top-1 accuracy, good Top-3 accuracy, and excellent Top-5 accuracy. This suggests that while pinpointing the single optimal measure remains challenging, model selections can fall within a near-optimal range. Specifically, Top-1 accuracy ranges from 25.5% (Grok 3) to 54.5% (Gemini 2.0); Top-3 accuracy ranges from 53.8% (Grok 3) to 86.3% (Gemini 2.0); and Top-5 accuracy ranges from 68.5% (DeepSeek R1) to 92.8% (ChatGPT o3). Top performers like ChatGPT o3 and Gemini 2.0 achieve over 45% at Top-1 accuracy, exceed 80% at Top-3 accuracy, and surpass 90% at Top-5 accuracy.

For payback year minimization, the best-performing LLM only reaches marginal accuracy even under the lenient Top-5 evaluation. This suggests that LLMs' ability to identify economically optimal retrofit measures remains limited. Specifically, Top-1 accuracy ranges from 6.5% (Gemini 2.0) to 14.3% (DeepSeek R1); Top-3 accuracy ranges from 15.0% (ChatGPT o1 and Grok 3) to

44.0% (Gemini 2.0); and Top-5 accuracy ranges from 19.5% (ChatGPT o1 and Grok 3) to 52.5% (Gemini 2.0). Gemini 2.0 is the only model to surpass 50% at Top-5 accuracy.

### 3.2. Consistency

#### 3.2.1. Overall agreement

Figure 5 shows the leave-one-out Fleiss' Kappa for each of the seven LLMs and overall Fleiss' Kappa for all models. It includes two types of overall agreement: (1) selection-based agreement (red bar), which measures agreement in the optimal retrofit selections recommended by the models; and (2) correctness-based agreement at the Top-1 (blue bar), Top-3 (orange bar), and Top-5 (green bar), which measures agreement whether models correctly identified the optimal retrofit within their top-$k$ results. The top chart shows these metrics for the objective of maximizing $CO_2$ reduction, while the bottom chart shows the same metrics for the objective of minimizing payback year. Bars labeled with "W/o" indicate the leave-one-out Fleiss' Kappa values when excluding one model at a time, while "Overall" represents the Fleiss' Kappa when all models are included. A higher leave-one-out Fleiss' Kappa value relative to the overall value suggests that the excluded model was lowering the agreement among the other models, whereas a lower leave-one-out value indicates that the excluded model made a positive contribution to the overall agreement.

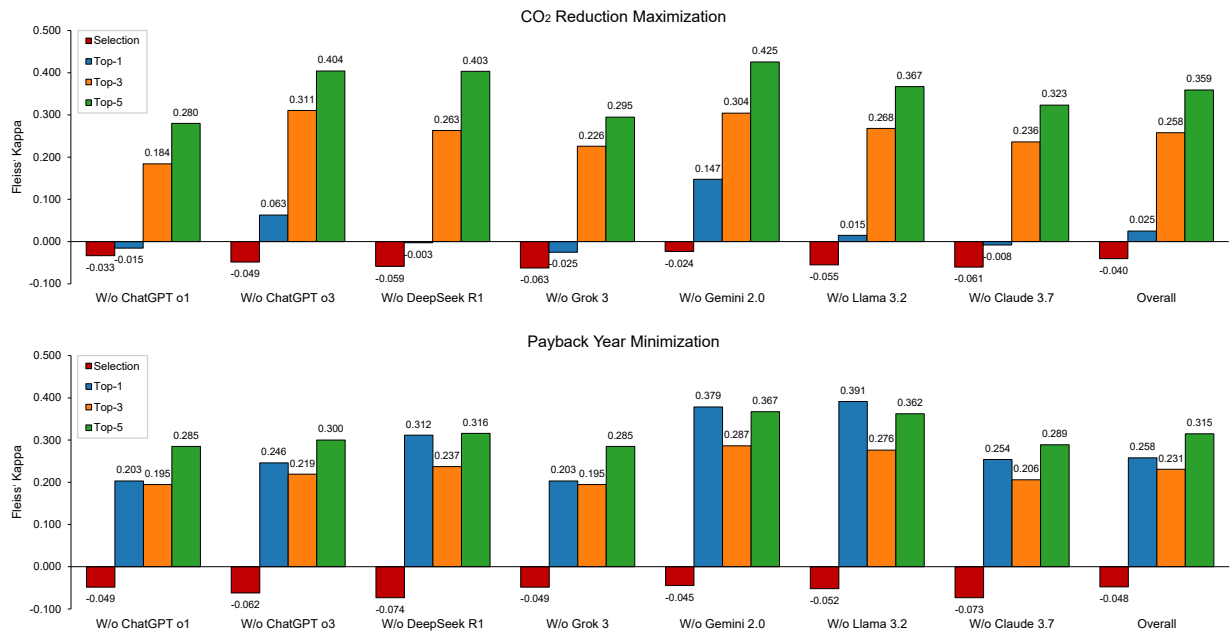

Figure 5. Overall agreement among LLMs in retrofit selection and correctness.

Across both objectives—$CO_2$ reduction maximization and payback year minimization—selection-based Fleiss' Kappa values remain consistently negative, indicating that the models' retrofit selections are less consistent than would be expected by chance. Correctness-based Fleiss' Kappa values fall within the fair agreement range at most. The overall Fleiss' Kappa for $CO_2$ reduction at Top-1 is particularly low (0.025), far below that for payback year (0.258). Correctness-based agreement at Top-3 and Top-5 is more comparable across the two objectives: for $CO_2$, the values are 0.258 and 0.359, while for payback year, they are 0.231 and 0.315.

For $CO_2$ reduction maximization, correctness-based Fleiss' Kappa values increase steadily from Top-1 to Top-5, starting from negative values (less than chance) and rising to as high as 0.425 (fair agreement). This trend aligns with the accuracy improvements shown in Figure 4 and suggests that LLMs show greater agreement in judging the correct optimal retrofit measures as accuracy improves. In addition, top performers in accuracy such as ChatGPT o3 and Gemini 2.0 also show the highest leave-one-out Fleiss' Kappa values (both larger than the overall), which indicates the high-accuracy models' correct judgments differ from others, thereby lowering overall agreement.

For payback year minimization, correctness-based Fleiss' Kappa values across Top-1, Top-3, and Top-5 fall within the fair agreement range (approximately 0.2 to 0.4). Fleiss' Kappa values peak at Top-1 (up to 0.391), decrease slightly at Top-3 (maximum 0.287), and rise at Top-5 (maximum 0.367). This trend, combined with the accuracy trend shown in Figure 4, reflects that at Top-1, most models perform poorly in a similar manner, leading to higher agreement; at Top-3, although all models exhibit improved accuracy, the divergence in the specific cases they correctly identify leads to a decline in agreement; and at Top-5, as more models converge in correctly identifying the same cases under more lenient criteria, agreement moderately increases.

*3.2.2. Pairwise agreement*

Figure 6 shows the pairwise Cohen's Kappa values among the seven LLMs for both objectives. These values complement the overall agreement shown in Figure 5 by providing pairwise agreement between models. Figure 6a and 6b show selection-based agreement; Figure 6c–e show correctness-based agreement at Top-1, Top-3, and Top-5 for $CO_2$ reduction; and Figure 6f–h show correctness-based agreement at corresponding Top-$k$ ranks for payback year. These comparisons provide a more detailed view of how the performance of individual models align with each other.

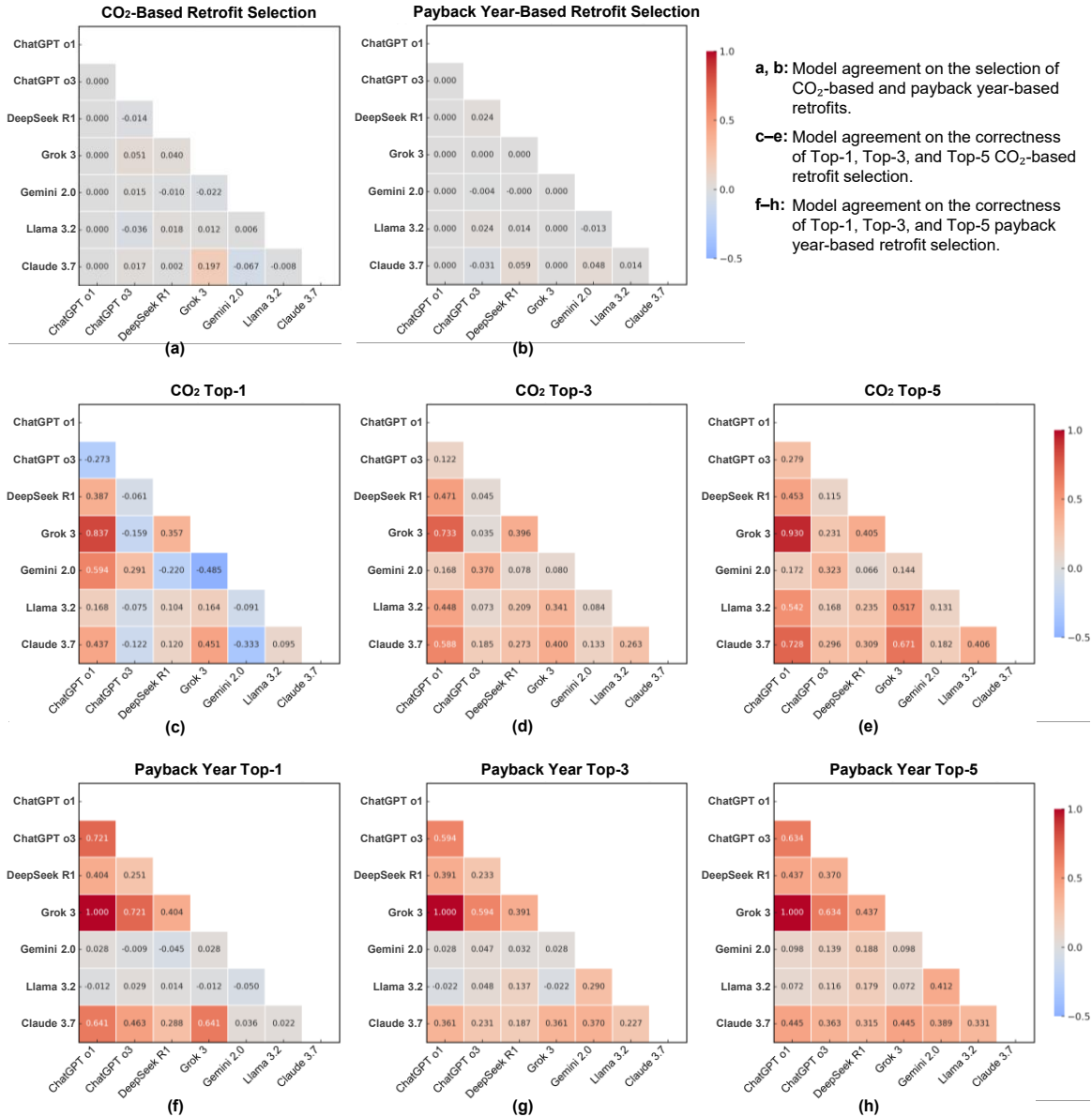

Figure 6. Pairwise agreement among LLMs in retrofit selection and correctness.

In selection-based agreement (Figure 6a and 6b), all Cohen's Kappa values are close to zero or negative, indicating low agreement in the retrofit selections by each pair of models—except for slight agreement observed between Grok 3 and Claude 3.7 for $CO_2$-based selection.

For $CO_2$ correctness-based agreement (Figure 6c–e), pairwise Cohen's Kappa values increase from Top-1 to Top-5, reinforcing the trend observed in Fleiss' Kappa (Figure 5). Beyond this, ChatGPT o3 and Gemini 2.0 consistently show the lowest agreement with other models in terms of correctness, further supporting the interpretation from Figure 5 that these two models tend to differ from others in their correctness judgments. Notably, ChatGPT o1 and Grok 3 demonstrate the highest pairwise agreement across all Top-*k* ranks, which is not captured by Fleiss' Kappa, suggesting that their correctness judgments are highly aligned throughout.

For payback year correctness-based agreement (Figure 6f–h), Cohen's Kappa values generally follow the same pattern as in Fleiss' Kappa, with a slight dip at Top-3 and a modest rise at Top-5. Additionally, Cohen's Kappa highlights Gemini 2.0 and Llama 3.2 having the weakest correctness-based agreement with other models, further reinforcing their distinct correctness profiles as indicated by the leave-one-out Fleiss' Kappa values (Figure 5). Meanwhile, ChatGPT o1 and Grok 3 again show perfect agreement from Top-1 to Top-5, providing additional evidence of their consistent alignment in recommending retrofit selections in same cases.

3.3 Sensitivity

Figure 7 presents the feature importance values for 28 input features in determining the optimal retrofit measures under two objectives: maximizing $CO_2$ reduction (left panel) and minimizing payback year (right panel). Each column corresponds to either the baseline, one of the seven LLMs, or the overall LLMs. In this analysis, feature importance represents how sensitive a model is to each feature—higher importance (represented by darker colors in the heatmap) indicates that the model relies more heavily on that feature when making retrofit decisions. Gray diagonal hatching indicates the LLM recommended the same retrofit selection for all 400 cases, making it impossible to compute feature importance due to a lack of variation in the outputs.

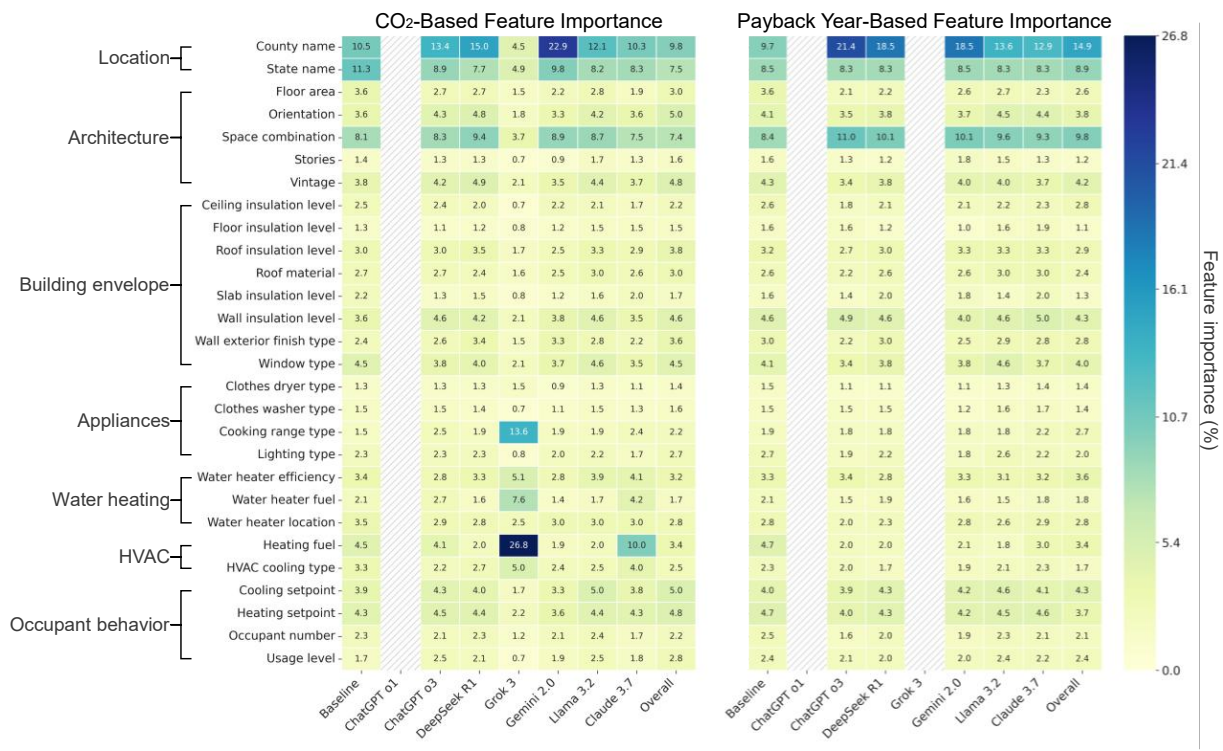

Figure 7. Feature importance comparison across LLMs for $CO_2$ and payback year-based retrofit decisions.

Across both $CO_2$ reduction maximization and payback year minimization, the patterns of feature importance are generally consistent between the baseline and the majority of LLMs. "County name", "State name", and "Space combination" consistently rank as the most important features, followed by a second tier that includes "Orientation", "Vintage", "Wall insulation level", "Window type", "Water heater efficiency", "Cooling setpoint", and "Heating setpoint". The remaining features are relatively less important in most cases, with only a few LLMs showing exceptions. These consistent sensitivity patterns suggest that most LLMs, like the baseline, prioritize location and architecture characteristics when selecting optimal retrofit packages.

While the general agreement is strong, some models exhibit notably different sensitivity patterns. Grok 3 shows an unusual focus on a few features for the $CO_2$ objective—mainly "Heating fuel", "Cooking range type", and "Water heater fuel". This focus deviates from both the baseline and other LLMs. It may help explain Grok 3's poor accuracy shown in Figure 4, where it ranks last at Top-1 and Top-3 and near the bottom at Top-5. Claude 3.7 also places relatively high weight on "Heating fuel", but its overall sensitivity pattern remains aligned with the baseline and most LLMs. In addition, most LLMs assign higher importance to "County name" than the baseline under both objectives. This suggests that LLMs rely heavily on building location features like "County name" to infer climate conditions, which in turn affect their judgment on energy use and retrofit selection.

3.4 Reasoning

Among the evaluated LLMs, ChatGPT o3 and DeepSeek R1 uniquely provide detailed, multi-step reasoning processes for selecting optimal retrofit packages. Their responses reflect a consistent five-step logic: (1) baseline assumption, (2) envelope impact adjustment, (3) system energy calculation, (4) appliance energy assumption, and (5) outcome comparison.

Figure 8 presents representative examples from ChatGPT o3 and DeepSeek R1 corresponding to each of these reasoning steps. While ChatGPT o3 presents its reasoning through a combination of explanatory text and embedded Python code, DeepSeek R1 conveys its logic entirely through descriptive narrative. Despite these differences in presentation, both models follow the same underlying logical progression. The multi-step logic aligns with engineering principles but remains simplified and considers only a narrow range of contextual dependencies.

| Reasoning Logic | ChatGPT o3 Reasoning Snapshot | DeepSeek R1 Reasoning Snapshot |
|---|---|---|
| **1. Baseline Assumption** The model estimates baseline energy loads based on building characteristics and location, such as cooling demand. | ```baseline['cooling_kwh']=1500
baseline['cooling_cost']=baseline['cooling_kwh']*elec_price
baseline['cooling_co2']=baseline['cooling_kwh']*co2_per_kwh``` | The heat pump capacity needed depends on the peak heating and cooling loads. But without detailed Manual J calculations, I'll estimate based on square footage. A rough estimate is 1 ton (12,000 BTU/h) per 400-500 sq ft. For 881 sq ft, maybe 2 tons (24,000 BTU/h) for cooling. Heating might be |
| **2. Envelope Impact Adjustment** The model adjusts heating and cooling loads based on assumed percentage reductions from envelope upgrades, such as insulation improvements and air sealing. | ```# adjust load for infiltration
if inf_insul:
    load*=0.85
    cooling_base*=0.85``` | If envelope reduces heating load by 20% (to 24 million BTU), then GSHP heating electricity: 24,000,000 / (4 * 3.412 * 1000) ≈ 24,000,000 / 13,648 ≈ 1,758 kWh. Cost: $210.96. Savings vs gas: 24 million BTU / (0.8 * 100,000) = 300 therms * $1.20 = $360. Savings: $360 - $210.96 = $149.04. |
| **3. System Energy Calculation** The model computes post-retrofit HVAC and water heating energy use by applying system efficiency metrics to cooling and heating loads, such as calculating electricity use from a heat pump's COP. | ```# heating energy
if i==4 or i==9 or i==14:
    # dual fuel: assume 70% heat supplied by he
    hp_elec=heating_elec_kwh(cop, load*0.7)
    gas_heat=load*0.3/0.8  # assuming same? act
    gas_heat_cost=gas_heat*gas_price_per_MMBTU
    gas_heat_co2=gas_heat*co2_per_MMBTU``` | GSHP has COP 4.0. Heating electricity: 30,000,000 / (COP * 3.412 * 1000) = 30,000,000 / (4 * 3.412 * 1000) ≈ 30,000,000 / 13,648 ≈ 2,198 kWh. Cost: $263.76. Savings vs gas: $450 - $263.76 = $186.24. |
| **4. Appliance Energy Assumption** The model assigns fixed annual energy use to upgraded appliances, such as for an upgraded cooking range. | ```# range
range_kwh = 80 if electrify else 100
range_cost=range_kwh*elec_price
range_co2=range_kwh*co2_per_kwh``` | Range: Electric resistance to induction. Assume 600 kWh/year saved. Savings: 600 * 0.12 = $72. |
| **5. Outcome Comparison** The model calculates total $CO_2$ reductions and payback years for each retrofit package, then compares results to identify the best option. | • Maximum CO₂ reduction → Package 13<br>  • Ultra-high-efficiency ASHP (SEER 24 / HSPF 13)<br>  • 30% air-sealing + R-60 attic<br>  • Full electrification (HP water heater, induction range, high-eff dryer)<br>  • ≈1.36tCO₂ saved per year (−43 % vs baseline)<br>• Fastest simple payback → Package 8<br>  • Same ultra-efficient ASHP & envelope upgrades without appliance electrification<br>  • Lower capital ($7.7k) but still captures the biggest space-conditioning savings<br>  • ≈40-yr payback — long, yet the best among the 16 options for this small home under current AL utility rates. | • Highest CO2 Reduction (Measure 15): This package combines a geothermal heat pump (GSHP) with full electrification of appliances, infiltration reduction, and insulation. It eliminates natural gas use for heating, cooling, and water heating, drastically reducing CO2 emissions. GSHPs are highly efficient, especially in heating, making this the most effective for CO2 savings despite higher upfront costs.<br>• Lowest Payback (Measure 16): Infiltration and insulation upgrades alone provide moderate energy savings at a low upfront cost. Improving the building envelope reduces heating/cooling loads without replacing major systems, resulting in a payback period of ~9 years, the shortest among all measures. |

Figure 8. Reasoning logic and illustrative snapshots from two LLMs.

The reasoning processes begin with a baseline assumption, where the models estimate a building's original energy load using basic characteristics such as floor area and location. This baseline serves as a reference for evaluating changes introduced by retrofits. Then, in the envelope impact adjustment step, the models account for the effects of building envelope retrofits, such as improved insulation or air sealing. These effects are typically represented by applying a percentage reduction to the baseline energy load, reflecting reduced heat loss. Next, during system energy calculation, the models adjust the updated energy consumption based on the efficiency of newly installed mechanical systems, such as heat pumps or furnaces. This step estimates energy consumption by dividing the energy demand by the system's efficiency. In the appliance energy assumption step, energy savings are estimated by comparing assumed consumption levels of appliances before and after retrofit. These assumptions reflect typical performance differences between appliance types and provide a basis for quantifying energy reductions. Finally, the outcome comparison step aggregates the estimated results to evaluate and prioritize retrofit packages. Although the reasoning procedures are similar across models, differences in assumed input values (such as baseline energy load and percentage energy reduction) lead to variation in estimated outcomes and final retrofit selections.

# 4. Discussion

The overall performance of LLMs in retrofit decision-making depends heavily on how effectively they process information at each stage of the workflow—prompt understanding, context representation, multi-step inference, and response generation. Weaknesses in any of these stages can result in oversimplified or incomplete decision-making. Identifying and addressing these limitations is essential to enhancing the performance of LLMs in building energy retrofit applications.

4.1. Prompt Input and Understanding

Effective prompt design is a foundational determinant of LLM performance. Studies have shown that even slight changes in prompt wording or structure can greatly influence LLMs' performance [37, 38]. We observed similar prompt sensitivity in our study. When prompted to *"identify the retrofit measure with the shortest payback period,"* most models selected the option with the lowest upfront investment, disregarding how energy cost savings can significantly influence payback outcomes. However, by simply adding the phrase *"considering both initial investment and energy cost savings"* to the same prompt, the models adjusted their reasoning to incorporate both factors. While the final recommendations were often unchanged, the revised prompt was more likely to trigger more comprehensive reasoning—highlighting that explicit guidance is necessary, as LLMs may not fully recognize the relevance of key variables unless directly prompted.

Prompt phrasing also influenced response generation. When asked to *"identify the measure with the greatest $CO_2$ reduction and the one with the shortest payback period,"* different models interpreted the task inconsistently: some returned only one option per objective, while others provided two or three. Notably, even for prompts implying a single optimal solution, such as *"the greatest"* or *"the shortest"*, some models still listed multiple alternatives. This suggests that LLMs do not always fully grasp the implicit expectations behind user prompts, and that clearer phrasing is needed to ensure alignment between user intent and model understanding.

These findings reinforce the importance of prompt engineering as a practical strategy for improving LLM performance in retrofit decision tasks. Unlike physics-based simulation tools, which rely on numerical inputs and fixed equations, LLMs generate responses based on the prompt and the knowledge they have learned from vast data. In this context, prompt plays a role similar to setting initial conditions in a simulation: providing clear parameters, objectives, and constraints helps guide the model toward more accurate results. Crafting effective prompts often requires iterative refinement and a solid understanding of how the model behaves in domain-specific contexts [39, 40]. By structuring prompts to include necessary building parameters, performance targets, and even hints for step-by-step reasoning, practitioners can significantly enhance the decision accuracy of LLM-generated retrofit advice. In short, prompt engineering emerges as a key enabler for harnessing LLMs in retrofit analyses, ensuring that these models remain grounded in context, as well-defined initial conditions steer a physics-based simulation toward reliable outcomes [41, 42].

## 4.2. Context Representation

After interpreting the prompt, LLMs internally form a contextual representation of the problem, essentially organizing the scenario, relevant variables, and their relationships in a way that supports reasoning. This stage is critical because it determines what information the model actively retains and uses when reasoning [43]. However, unlike physics-based models that explicitly define all variable interdependencies through equations, LLMs construct this representation using language-based associations. As a result, some contextual elements, especially those with lower salience, may be excluded from context representation when their necessity is not explicitly stated.

Although sensitivity analysis shows that most LLMs assign similar importance to input features as the physics-based baseline, many of these features are not effectively integrated into the models' context representation and thus are not available to be used in reasoning. For example, "usage level" consistently received a moderate importance—around 2.5% on average across models—yet played no role in the reasoning leading to the final retrofit recommendation. This suggests that while LLMs can identify which features are generally relevant, they often fail to embed them into a coherent contextual understanding of the task. Unlike physics-based models that systematically account for all input variables, LLMs tend to form context representations around only a few salient cues. As a result, even when feature importance aligns with the baseline, an incomplete or fragmented context representation can limit the reasoning process and reduce overall accuracy.

Improving context representation requires not only well-structured prompts, but also strategies that explicitly indicate the necessity of certain input features. This enables LLMs to incorporate features that may appear less important relative to others but are still essential for accurate decision-making. [44]. In our study, although the prompts included relevant information such as occupancy levels, we did not emphasize the necessity of these variables within the prompt. As a result, models often overlooked them during reasoning. If the prompts had more clearly directed the model's attention—for example, by stating that certain features directly impact energy performance—LLMs may have been more likely to incorporate them into their reasoning and generate more accurate recommendations.

## 4.3. Inference and Response Generation

The inference and response generation of LLMs exhibits three major limitations: oversimplified decision logic, inconsistency across responses, and unintended context-driven bias. These issues, while not always visible from the final outputs, were clear through the step-by-step reasoning traces.

First, models appeared to rely on simplified reasoning patterns in our study. For instance, assuming that the retrofit measure with the lowest upfront cost would always yield the shortest payback period, or that fossil-fuel systems should be automatically replaced by electric alternatives. These rules were applied uncritically without evaluating performance trade-offs. To address this, prompts can be refined to explicitly emphasize multi-criteria trade-offs, such as balancing upfront investment with long-term energy saving. Incorporating counterexamples into prompts may also discourage overgeneralization. Additionally, providing structured reasoning templates, such as listing each step for evaluating cost, energy saving, and payback, can help enforce more nuanced inference chains.

Second, we observed that LLM-generated results exhibited a degree of variability and uncertainty that is atypical of physics-based methods. Because LLMs generate text probabilistically, repeated queries or slight rephrasing can yield inconsistent answers [45, 46]. This may lead to conflicting retrofit recommendations, even when the underlying input remains the same. Such inconsistency limits the suitability of LLMs for high-precision, reliability-critical tasks [47, 48], such as hourly energy consumption prediction. To address these challenges, several strategies can be employed to improve the reliability of LLM applications. First, fine-tuning LLMs with domain-specific datasets can reduce variability and uncertainty [49, 50]. Second, domain-specific small language models distilled from large LLMs can provide faster and more consistent inference while retaining essential reasoning capabilities [51, 52]. Third, retrieval-augmented generation can ensure responses are grounded in validated sources, improving consistency in retrofit recommendations [53, 54]. Finally, hybrid modeling can leverage LLMs for interpreting inputs and generating hypotheses, while physics-based simulations validate results, ensuring traceable and repeatable outcomes [24, 55].

Third, in dialogue-based LLM interactions, context carryover may introduce unintended bias when similar but independent queries are asked sequentially. Because the model retains elements of prior responses, subsequent answers may be influenced by earlier reasoning, even if each query deserves separate consideration [56, 57]. This undermines reliability and makes it difficult to compare responses across different scenarios. To counter these effects, users can reset or isolate the conversation context, rephrase questions to stand alone, or provide explicit directives instructing the model to ignore previous content. Conversely, a structured chain-of-thought approach can turn sequential prompting into an advantage. By guiding the model step by step, complex tasks are broken into smaller logical steps. Each step can be checked or refined before moving on, resulting in more coherent responses. Users can reinforce this by requesting intermediate reasoning, introducing checkpoints to correct errors, and encouraging the model to "think aloud." Thus, chain-of-thought prompting leverages the model's sequential nature for clarity and accuracy, rather than allowing context to create unwanted bias [58, 59].

## 5. Conclusion

This study explores the potential of generative AI to address key challenges of building energy retrofit decision-making, including poor generalizability and low interpretability. These limitations often hinder the effectiveness and real-world adoption of conventional retrofitting applications. In this work, seven leading LLMs are tasked with generating energy retrofit decisions under two distinct contexts: a technical context focused on maximum $CO_2$ reduction, and a sociotechnical context focused on minimum payback period. The AI-generated retrofit decisions are evaluated using three criteria: whether the recommendation matches the top-ranked retrofit measure (Top-1), falls within the top three (Top-3), or within the top five (Top-5). Model performance is evaluated across four dimensions—accuracy, consistency, sensitivity, and reasoning—using a sample of 400 diverse homes from the ResStock 2024.2 dataset, spanning 49 U.S. states.

Table 2. Summary of evaluation results.

|  | **Technical Context** | **Sociotechnical Context** |
|---|---|---|
| **Accuracy** | An upward trend is observed (from 38.8% to 79.0%) as the criteria become less stringent (from Top-1 to Top-5). AI decisions can match top-1 retrofit measures up to 54.5% of cases and fall within the top-5 measures up to 92.8% of cases. | An upward trend is observed (from 11.1% to 32.6%), as the criteria become less stringent (from Top-1 to Top-5). AI decisions can match top-1 retrofit measures up to 14.3% of cases and fall within the top-5 measures up to 52.5% of cases. |
| **Consistency** | A sharp upward trend is observed (from Kappa 0.025 to 0.359) when the criteria become less stringent (from Top-1 to Top-5). | Almost no trend is observed (from Kappa 0.258 to 0.315) when the criteria become less stringent (from Top-1 to Top-5). |
| **Sensitivity** | Very sensitive to climate-related factors, e.g., location. Sensitive to building size and space combination. | Very sensitive to climate-related factors, e.g., location. Sensitive to building size and space combination. |
| **Reasoning** | A five-step reasoning workflow is found. | A five-step reasoning workflow is found. |

The evaluation results, summarized in Table 2, demonstrate that LLMs are capable to produce effective retrofit decisions, although struggling to pinpoint the best one. Their performance is notably stronger in identifying retrofit packages in the technical context (to maximize $CO_2$ reduction) than in the sociotechnical context (to minimize payback year). This difference likely reflects the relative clarity and consistency of technical optimization objectives, which are more easily captured by model reasoning, whereas sociotechnical considerations involve trade-offs between economic, behavioral, and contextual factors that may be more difficult for LLMs to interpret and prioritize accurately. While top-performing models such as ChatGPT o3 and Gemini 2.0 achieved high accuracy for $CO_2$-based decisions, all models exhibited limited effectiveness in selecting economically optimal retrofits. Noteworthy is that their accuracy can reach as high as 54.5% in matching Top-1 and 92.8% in matching Top-5 without model fine-tuning. Consistency analysis revealed low overall agreement among models, with higher-accuracy models more likely to diverge from others and lower-accuracy models showing greater alignment. Sensitivity analysis showed that most LLMs prioritized similar features as the physics-based baseline—mainly location and architecture. However, Grok 3 deviated from this pattern, assigning disproportionate importance to a few features, which may explain its lower accuracy. The reasoning processes of ChatGPT o3 and DeepSeek R1 reflected structured, multi-step logic aligned with engineering principles. However, their reasoning remained simplified and omitted necessary contextual dependencies. Overall, while the decision logic and feature focus of LLMs are generally reasonable, their accuracy, consistency, and contextual understanding must be improved before they can be reliably applied to retrofit decision-making in practice.

**CRediT authorship contribution statement**

**Lei Shu:** Conceptualization, Methodology, Investigation, Data curation, Formal analysis, Visualization, Writing – original draft, Writing – review & editing. **Dong Zhao:** Conceptualization, Methodology, Formal analysis, Writing – review & editing, Supervision, Funding acquisition.

**Acknowledgements**

This study was supported by the National Science Foundation (NSF) of the United States through Grant #2046374. Any opinions, findings, conclusions, or recommendations expressed in this material are those of the researchers and do not necessarily reflect the views of NSF.